\documentclass{SCIS2026}
\usepackage[ruled, vlined]{algorithm2e}
\usepackage{makecell}
\usepackage{epstopdf}
\usepackage{breakurl}
\usepackage{lineno}

\newcommand{\upmodels}{{\perp \!\!\! \perp}}

\usepackage{soul}
\usepackage{color, xcolor}
\definecolor{lightblue}{RGB}{0, 191, 255}
\sethlcolor{lightblue}
\soulregister\cite7
\soulregister\eqref7
\soulregister\ref7

\begin{document}
\ArticleType{REVIEW}
\Year{2025}
\Month{January}
\Vol{68}
\No{1}
\DOI{}
\ArtNo{}
\ReceiveDate{}
\ReviseDate{}
\AcceptDate{}
\OnlineDate{}
\AuthorMark{}
\AuthorCitation{}

\title{Spatio-Temporal Graphical Counterfactuals: \\ An Overview}{Kang M Y, Chen D X, Pu Z Y, et al. Spatio-Temporal Graphical Counterfactuals: An Overview}


\author[1]{Mingyu Kang}{}
\author[2]{Duxin Chen}{{chendx@seu.edu.cn}}
\author[3]{Ziyuan Pu}{}
\author[4, 5]{Jianxi Gao}{}
\author[2, 6, 7]{Wenwu Yu}{{wwyu@seu.edu.cn}}


\address[1]{School of Cyber Science and Engineering, Southeast University, Nanjing {\rm 210096}, China}
\address[2]{School of Mathematics, Southeast University, Nanjing {\rm 210096}, China}
\address[3]{School of Transportation, Southeast University, Nanjing {\rm 210096}, China}
\address[4]{Department of Computer Science, Rensselaer Polytechnic Institute, Troy NY {\rm 12180}, USA}
\address[5]{Center for Network Science and Technology, Troy NY {\rm 12180}, USA}
\address[6]{Frontiers Science Center for Mobile Information Communication and Security, Nanjing {\rm 210096}, China}
\address[7]{Purple Mountain Laboratories, Nanjing {\rm 211102}, China}

\abstract{

Counterfactual thinking is a crucial yet challenging topic for artificial intelligence to learn knowledge from data and ultimately improve performance for new scenarios. Many research works, including the Potential Outcome Model (POM) and the Structural Causal Model (SCM), have been proposed to address this. However, their modeling, theoretical foundations, and application approaches often differ. Moreover, there is a lack of graphical approaches for inferring spatio-temporal counterfactuals, that account for spatial and temporal interactions among multiple units. Thus, in this work, we aim to present a survey that compares and discusses different counterfactual models, theories and approaches. Additionally, we propose a unified graphical causal framework to infer spatio-temporal counterfactuals.

}


\maketitle


\section{Introduction}

How can we enable an intelligent machine to think about and answer a spatio-temporal counterfactual question? For example, sometimes we want to know: if a different investment strategy had been implemented, would we have obtained higher returns? Or, in computer networks, what changes would occur in network load if a node's configuration had never been changed? Or would someone still purchase a product even if they had never seen the advertisement? These questions generally follow a spatio-temporal counterfactual pattern: ``Given the observed $\dots$, if $\dots$ had been done, how would $\dots$ have evolved?''. Thus, according to the pattern, time must be considered first, because the questions are posed now, but the counterfactual actions are supposed to be performed in the past. Second, counterfactual outcomes are unobservable. The outcomes of some actions are observed when they are performed on a real-world system, thus, these outcomes are factual. But the counterfactual outcomes are not observed, and they are imagined by supposing that different actions had been taken. Third, the real-world system can be observed as a multivariate time series, and there are widespread multivariate interactions within it~\cite{Era2024, Pravicy2025}.

To enable intelligent machines to mimic counterfactual thinking, the Potential Outcome Model (POM)~\cite{Causality2015rubin} and Structural Causal Model (SCM)~\cite{Causality2009, APrimer2016, Elements2017, why2018pearl} have been proposed as two frameworks with different foundations. POM uses stable unit treatment value assumption (SUTVA), consistency assumption, positivity assumption and ignorability assumption as foundations~\cite{consistency1976rubin, consistency1978rubin, SUTVA1980rubin, PSM1983rubin}. If these assumptions are hold, one can use matching or imputation approaches to infer counterfactuals without confounders. In contrast, SCM builds a causal ladder, i.e., abduction, action and prediction, to infer counterfactuals through graphical language d-separation~\cite{Causality2009, APrimer2016, why2018pearl}. With the graphical language d-separation, the causal Markov assumption, faithfulness assumption and causal sufficiency assumption are proposed to guarantee the correctness of counterfactual inference on SCM~\cite{BayesianNetwork1985pearl, belief1986pearl, simulation1987pearl, IntelligentSystem1988pearl, IdentifyDuLi1990pearl}.

However, the distinct foundations between POM and SCM lead to differences in counterfactual inference. These differences include: (i) POM assumes positivity, whereas SCM does not. Thus, the counterfactuals in POM are factual and can be found in observations certainly. In contrast, counterfactuals in SCM can be out of observations and are therefore unfalsifiable, that is, their correctness cannot be empirically validated. (ii) POM explicitly assumes consistency, but this assumption is rejected for SCM~\cite{why2018pearl}. Actually, SCM also implicitly relies on the consistency assumption to ensure that the counterfactual outcome corresponds to one of the possibly unobserved potential outcomes. (iii) SCM uses a type of graphical language to deconfound with certain rules, but POM mainly relies on human-guided deconfounding, which makes it difficult to satisfy the ignorability. In contrast, SCM can deconfound via the back-door criterion~\cite{diagram1995pearl, overview2009pearl, Causality2009, APrimer2016} under the casual sufficiency assumption, and it is equivalent to ignorability. (iv) POM faces limitations to analyze indirect causation (mediation), thus POM typically relies on the Baron-Kenny model~\cite{BK1986, Mediation2022heyas} for assistance. By contrast, SCM facilitates this analysis through the front-door criterion~\cite{diagram1995pearl, overview2009pearl, Causality2009, APrimer2016}, which decomposes a causal pathway into two segments: from the source variable to mediator, and from the mediator to target variable. Back-door adjustment is then applied twice, assuming no additional mediators are presented.

Moreover, both POM and SCM currently lack the ability to model the spatio-temporal interactions among system units. For example, the SUTVA requires no interaction between system units in POM. Similarly, SCM, when based on the traditional Bayesian network, does not handle multivariate time series and often overlooks nonstationarity. This limitation motivates us to identify a common point to unify the two frameworks and strengthen inference for spatio-temporal counterfactuals. Our recent work~\cite{UCN2022, NHCE2024} has proposed a spatio-temporal Bayesian networks (STBNs) model based on a temporal assumption that the cause precedes or parallels the effect in time. It is found that the famous complex networks~\cite{resilience2016jianxigao, identification2022tpi, identification2022tsp}, a type of network dynamics used to model spatio-temporal interactions, are also a special case of STBNs.

\begin{figure}[htbp]
	\centering
	\includegraphics[width=\linewidth]{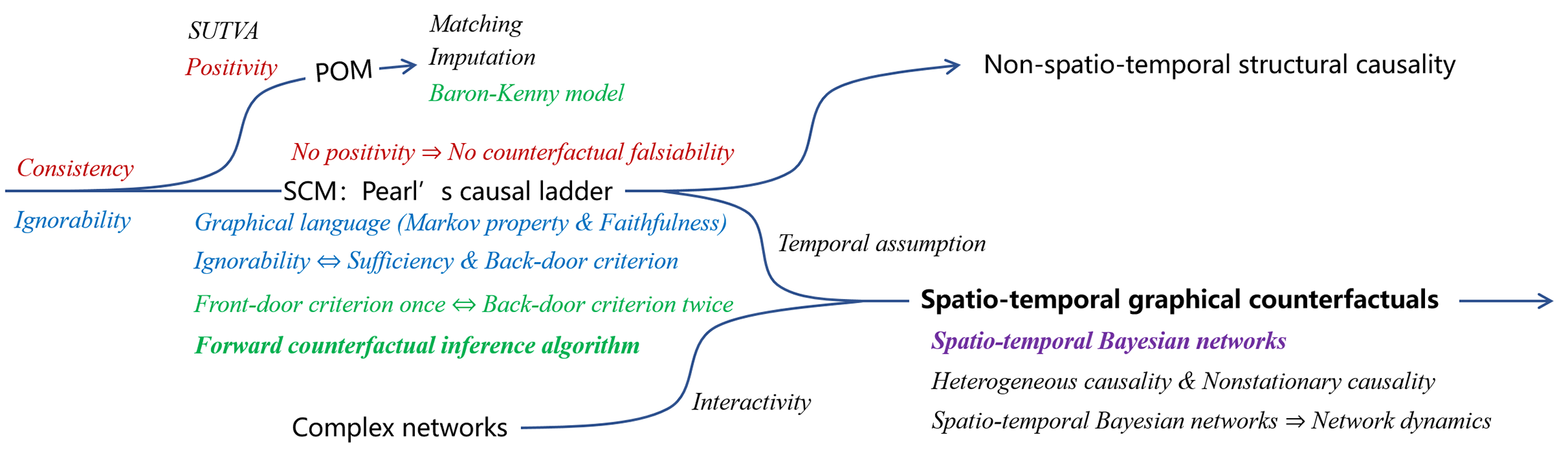}
	\caption{Towards spatio-temporal graphical counterfactuals.}
	\label{figure: storyline}
\end{figure}

Thus, we further conduct this work to build a spatio-temporal graphical counterfactual framework based on STBNs. This framework requires two components: (i) STBNs used as the foundation of SCM, (ii) a Forward Counterfactual Inference Algorithm designed on the decomposability of the front-door criterion. We then discuss how to get the two points, following the storyline as shown in Fig.~\ref{figure: storyline}. Thus, the main contributions of this work are as follows:
\begin{enumerate}
	\item[1.] A survey is conducted to discuss the frameworks from POM to SCM, including their theoretical foundations and application approaches. This part is organized from Section~\ref{section: pom} to Section~\ref{section: diffs}.
	
	\item[2.] A Forward Counterfactual Inference algorithm is designed here to recursively and autonomously infer counterfactuals through causal graphical languages. This part is presented as \textbf{Algorithm~\ref{algorithm: FCI}} in Section~\ref{section: fci}.
	
	\item[3.] An overview of the spatio-temporal graphical counterfactual framework is proposed to discuss the spatio-temporal Bayesian networks, their nonstationarity, and the relationship with complex networks. This part is organized in Section~\ref{section: stgc}.    	
\end{enumerate}

\section{Potential Outcome Model}
\label{section: pom}

The counterfactuals are defined as the unobserved potential outcomes in POM, as shown in Table~\ref{table: pom}. Let $Y$ be the potential outcomes of the target variable, and $T$ is the treatment variable. $T=1$ represents assignment to the treatment group, and $T=0$ represents assignment to the control group in a randomized controlled trial involving $N$ units. $X$ represents covariates that have relationship with $Y$ and $T$, and are controlled for randomized trials. POM models the general process from causal estimands ($T$, $Y$ and $X$) to ``Science''~\cite{POM2005}. In this case, for different treatments $T$, only one potential outcome, $Y(1)$ or $Y(0)$, can be observed, and another one is unobserved, namely the counterfactual.

\begin{table}[htbp]
	\centering
	\caption{Data for POM example.}
	\label{table: pom}
	\renewcommand\arraystretch{1.3}
	\setlength{\tabcolsep}{6.0mm}
	\small
	\begin{tabular}{ccccc} 
		\hline 
		Units & Covariates  & \multicolumn{2}{c}{Potential outcomes} & Treatment\\
		\cline{3-4}
		$i$ & $X$ & $Y(1)$ & $Y(0)$ & $T$\\
		\hline  
		1 & $x_1$ & $y_1(1)$ & $?$ & $1$\\
		2 & $x_2$ & $?$ & $y_2(0)$ & $0$ \\
		3 & $x_3$ & $?$ & $y_3(0)$ & $0$\\
		4 & $x_4$ & $y_4(1)$ & $?$ & $1$\\
		5 & $x_5$ & $y_5(1)$ & $?$ & $1$\\
		$\vdots$ & $\vdots$ & $\vdots$ & $\vdots$ & $\vdots$ \\
		$N$ & $x_N$ & $?$ & $y_N(0)$ & $0$\\
		\hline 
	\end{tabular}
\end{table}

\subsection{Inferring Counterfactuals through POM}
\label{section: app_pom}

To infer the counterfactuals using POM, one common approach is exact sample matching. For a unit $i$, one can search the samples with exactly matching covariates $X$ but treatment $T$ is different, and then estimate the counterfactual outcome $Y_i(1-T_i)$. But it is difficult to conduct when the dimension of $X$ is high, as it requires a large number of samples to support the matching. Thus, approximate matching is commonly better, e.g., caliper matching~\cite{caliper1970rubin, caliper1973rubin, caliper1976rubin, caliper2008rubin} and propensity score matching~\cite{PSM1983rubin, PSM1996rubin, PSM1997rubin, PSM2008}. They search for a group of similar samples to the target unit with one or more measurements, then calculate counterfactuals.

Another common approach is data-driven imputation, which views counterfactuals as missing values, and interpolates them by fitting the observational data. Linear or nonlinear regressions~\cite{LR1901pearson, logistic1958, var1980, VARX1988} are typically used to interpolate missing values based on patterns learned from the observational data. In contrast, tensor decomposition approaches~\cite{TensorD2009SIAM, TensorD2011, recommender2016} recover the missing values by decomposing and reconstructing a sparse data tensor. Deep generative models~\cite{VAE2013, GAN2014, FLOW2015, DDPM2020, DDIM2021, ScoreBased2021ICLR, KMY2024TPS} have also been used to capture data uncertainty and generate missing values by randomly sampling. Note that, the imputed missing values produced by generative models are not unique. Instead, they constitute a distribution of plausible values, which distinguishes them from regression-based and tensor decomposition approaches.

\subsection{Foundational Assumptions of POM}

Before using the above approaches within POM framework, four assumptions must be accepted. The first is SUTVA, which is defined as 
\begin{assumption}[SUTVA~\cite{SUTVA1980rubin}]
	In Table~\ref{table: pom}, 
	\begin{enumerate}
		\item[1.] There is no interference between units, that is, neither $Y_i(1)$ nor $Y_i(0)$ is affected by what action any other unit received.
		\item[2.] There are no hidden versions of treatments, that is, no matter how unit $i$ received treatment $T=1$, the outcome that would be observed would be $Y_i(1)$, and similarly for treatment $T=0$. 
	\end{enumerate}
	\label{assumption: sutva}
\end{assumption}
SUTVA guarantees the identical independence between the experimental units, and for unit $i$, the outcome $Y_i$ is only up to its treatment $T_i$, not the others'. However, this assumption is not always satisfied, e.g., in social networks, where interactions occur at varying times. 

The second assumption is consistency, defined as
\begin{assumption}[Consistency~\cite{consistency1976rubin, consistency1978rubin}]
	In Table~\ref{table: pom}, if the $i$-th unit is selected for a treatment $T_i$, the observed value of $Y_i$, neither $ Y_i(1)$ nor $Y_i(0)$, is the same for all assignments of treatments to the other experimental units. 
	\label{assumption: consity}
\end{assumption}
Another form to describe the consistency assumption through expectation $\mathbb{E}[\cdot]$ is
\begin{equation}
	\begin{aligned}
		&\mathbb{E}[Y(1)|T=1] = \sum_{i=1}^N y_i \times t_i = \mathbb{E}[Y|T=1], \\
		&\mathbb{E}[Y(0)|T=0] = \sum_{i=1}^N y_i \times (1 - t_i) = \mathbb{E}[Y|T=0],
	\end{aligned}
\end{equation}
where $y_1, \dots, y_N$ are the samples for $N$ units, and $t_1, \dots, t_N \in \{0, 1\}$ are the values of treatment variable $T$, as shown in Table~\ref{table: pom}. This is the same as the case of taking a conditional expectation with respect to $X$. Thus, the consistency assumption guarantees that the missing counterfactual values in Section~\ref{section: app_pom} can be interpolated by the other observational values.

The third assumption is positivity, defined as
\begin{assumption}[Positivity~\cite{PSM1983rubin}]
	In Table~\ref{table: pom}, $0 < P(T=1| X=x) < 1$.
	\label{assumption: posity}
\end{assumption}
This means that, for $X=x$, some treatments are always assigned to $T=0$ or $T=1$ at random. Otherwise, the counterfactuals for $T=1$ could not be obtained if all units with $X=x$ are assigned with $T=0$, and this is the same as the counterfactuals for $T=0$.

The fourth assumption is ignorability, defined as  
\begin{assumption}[Ignorability~\cite{consistency1976rubin, consistency1978rubin}]
	In Table~\ref{table: pom}, $T \upmodels Y(1), Y(0) | X$ for each unit.
	\label{assumption: igno}
\end{assumption} 
Thus, the ignorability assumption implies that all the confounders have been observed in $X$, that is, no hidden confounder would interfere with the observations of $Y(1), Y(0)$. Thus, the ignorability assumption is also called unconfoundedness assumption. This assumption is usually made to avoid the preference for experimental manipulation, which would guarantee randomization in the experiments~\cite{randomization1974rubin}.

\section{Structural Causal Model}

In contrast to POM, SCM provides a graphical language for intuitively discussing causality, thereby avoiding complex human-guided technical operations. An SCM is built on a set of variables, including endogenous variables $\mathbf{V} = \{X_1, \dots, X_n\}$ and exogenous variables $\mathbf{U}=\{U_1, \dots, U_n\}$. Note that, endogenous variables are observable from data, but exogenous variables are not. Moreover, a set of functions $\mathbf{F}=\{f_1, \dots, f_n\}$ can also be used to describe the functional relationship between these variables. 

\begin{figure}[htbp]
	\centering
	\begin{minipage}[b]{0.43\textwidth}
		\centering
		\includegraphics[width=\linewidth]{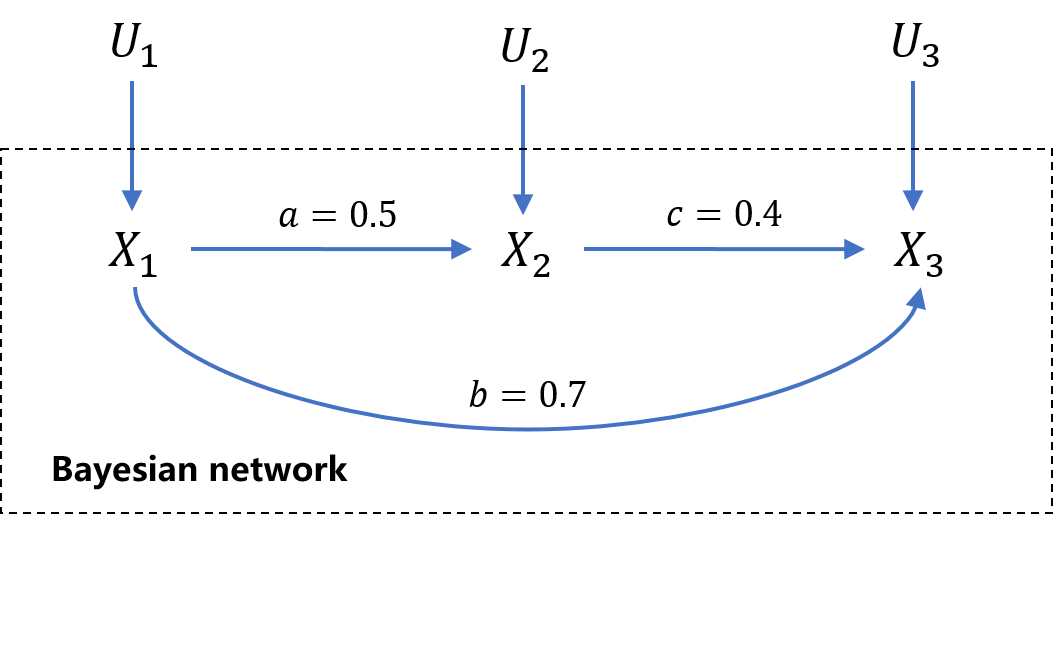}
		\caption{Example of SCM~\cite{APrimer2016}.}
		\label{figure: scm}
	\end{minipage}
	\hfill
	\begin{minipage}[b]{0.56\textwidth}
		\centering
		\includegraphics[width=\linewidth]{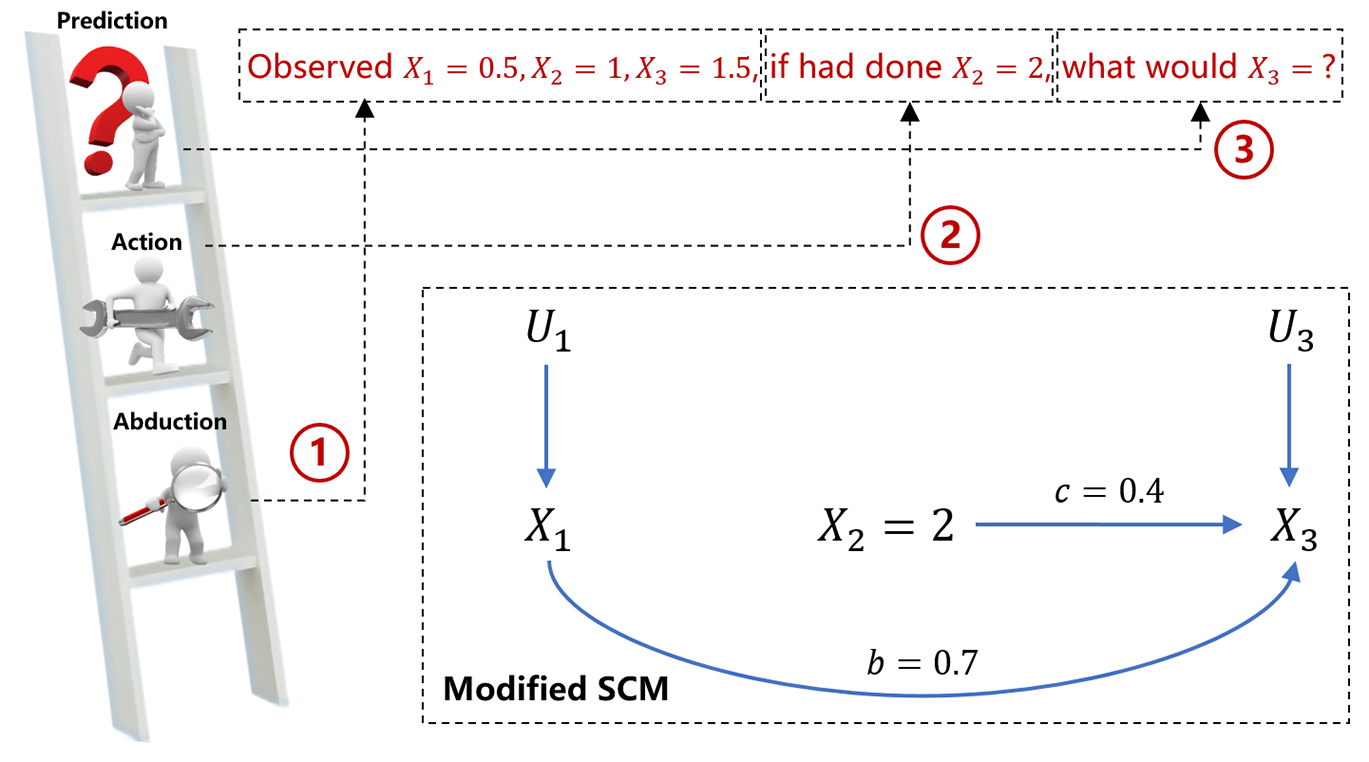}
		\caption{Diagram of Pearl's causal ladder~\cite{APrimer2016}.}
		\label{figure: ladder}
	\end{minipage}
\end{figure}

For example, as shown in Fig.~\ref{figure: scm}, the SCM can be functionally described as 
\begin{equation}
	\begin{aligned}
		f_1:& X_1 = U_1, \\
		f_2:& X_2 = aX_1 + U_2, \\
		f_3:& X_3 = bX_1 + cX_2 + U_3,
	\end{aligned}
\end{equation}
where $a=0.5, b=0.7, c=0.4$. $U_1, U_2, U_3$ are all additive noise, and they are mutually independent.

In an SCM, the functional relationship must conform to the directed acyclic graphical constraint. The graph is originally based on the concept of Bayesian network~\cite{BayesianNetwork1985pearl, belief1986pearl, simulation1987pearl, IntelligentSystem1988pearl, IdentifyDuLi1990pearl}. And later, it is also called as causal graphical model to highlight its causal structure~\cite{graphical1993pearl, diagram1995pearl, overview2009pearl, Causality2009, APrimer2016}. And recently, it is also called as causal network in the field of causal discovery~\cite{CD2019glymour, CD2021bernhard, CD2022DAG, CD2022JAIR, CD2023timeseries}. But actually, they are the same in the framework of SCM, and they are all directed acyclic graph (DAG) that represents Markovian knowledge. Thus, to avoid misunderstanding, we here use the name of Bayesian network uniformly, due to the fact that causal graphical model and causal network are not always a Bayesian network in Fig.~\ref{figure: scm}. For example, direct cyclic graphs~\cite{DCG2008peter, DCG2013peter, DCG2017mooij, DCG2021mooij}, Markov networks~\cite{PGM2009, MarkovNetwork2018}, and full time graph~\cite{Elements2017}, they are also causal graph or causal network.

\subsection{Pearl's Causal Ladder~\cite{Causality2009, APrimer2016, why2018pearl}}
\label{section: pcl}

To infer the counterfactuals through SCM, there are three steps:
\begin{enumerate}
	\item[1.] \textbf{Abduction:} Infer the values of exogenous variables $\mathbf{U}$ from the observational data;
	\item[2.] \textbf{Action:} Perform an intervention on SCM, e.g., $do(X_2=2)$ on SCM in Fig.~\ref{figure: scm}, and then, $X_2$ would be assigned with value $2$, and the arrows to $X_2$ would be modified, as shown in Fig.~\ref{figure: ladder};
	\item[3.] \textbf{Prediction:} Use the modified SCM to recalculate the counterfactuals of the target variable. 
\end{enumerate}

As shown in Fig.~\ref{figure: ladder}, consider the following query ``observed $X_1=0.5, X_2=1, X_3=1.5$, if had done $X_2=2$, what would $X_3 = ?$''. Obviously, this is a counterfactual question. Thus, following Pearl's causal ladder, as shown in Fig.~\ref{figure: ladder}, we first infer the values of exogenous variables $\mathbf{U}=\{U_1, U_2, U_3\}$ from the observational data $X_1=0.5, X_2=1$ and $X_3=1.5$, as follows:
\begin{equation}
	\begin{aligned}
		U_1 &= 0.5, \\
		U_2 &= 1 - 0.5 \times 0.5 = 0.75, \\
		U_3 &= 1.5 - 0.7 \times 0.5 - 0.4 \times 1 = 0.75. 
	\end{aligned}
\end{equation}
Then, $do(X_2 = 2)$ to obtain a modified SCM. And finally, recalculate the counterfactual of $X_3$, that is, 
\begin{equation}
	\begin{aligned}
		X_3(do(X_2 = 2)) = 0.5 \times 0.7 + 2 \times 0.4 + 0.75 = 1.9,
	\end{aligned}
\end{equation}
where $X_3(do(X_2 = 2))$ is different compared to the observational $X_3$. 

The intervention, also named as $do$-operator~\cite{Causality2009}, can be generalized into the form of probability, if the modularity assumption is introduced as follows:
\begin{assumption}[Modularity~\cite{Causality2009}]
	If a set of variables $\{X_{j_1}, \dots, X_{j_p}\} \subset \mathbf{V}$ is intervened, then for each variable $X\in \mathbf{V}$, it is obtained that
	\begin{enumerate}
		\item[1.] if $X\notin \{X_{j_1}, \dots, X_{j_p}\}$, then $P(X| \mathbf{Pa}(X))$ remains unchanged. Here, $\mathbf{Pa}(X)$ denotes the set of predecessors of $X$ in a Bayesian network, also called the causal parents of $X$.
		
		\item[2.] if $X \in \{X_{j_1}, \dots, X_{j_p}\}$, then $P(X = x| \mathbf{Pa}(X)) = 1$ if $x$ is the value set by intervention to $X$, otherwise, $P(X = x| \mathbf{Pa}(X)) = 0$. \\
	\end{enumerate}
\end{assumption} 
This means that if a set of interventions is applied to $X_{j_1}, \dots, X_{j_p}$, their values would be fixed. The connections between them and their causal parents would be broken, and their causal parents would not affect them anymore in the modified SCM, just like in Fig.~\ref{figure: ladder}. Moreover, this also means that the probability distributions of the other variables that are not intervened upon would not change. 

\subsection{Foundational Assumptions of Bayesian Network}

As presented above, to clearly define causality, each SCM is associated with a Bayesian network satisfying the directed acyclic graphical constraint. Thus, the causality in the Bayesian network provides the foundation for counterfactuals. Thus, we discuss the foundational assumptions of Bayesian networks in the following sections. 

First of all, a concept of d-separation can be defined on a Bayesian network $\mathcal{G}$ as
\begin{definition}[d-separation~\cite{BayesianNetwork1985pearl, belief1986pearl, simulation1987pearl, IntelligentSystem1988pearl, IdentifyDuLi1990pearl}]
	Let $\mathbf{X}$, $\mathbf{Y}$ and $\mathbf{Z}$ be three disjoint subsets of endogenous variables $\mathbf{V} = \{X_1, \dots, X_n\}$, and let $\mathbf{p}$ be any path from a node in $\mathbf{X}$ to a node in $\mathbf{Y}$ regardless of direction. $\mathbf{Z}$ is said to block $\mathbf{p}$ if and only if there is a node $v\in \mathbf{p}$ satisfying one of the following items:
	\begin{enumerate}
		\item[1.] $v$ has {\rm v}-structure (two nodes $a, b\in \mathbf{p}$ pointing to $v$, namely $a\rightarrow v\leftarrow b$), and neither $v$ nor its any descendants are in $\mathbf{Z}$; 
		\item[2.] $v$ in $\mathbf{Z}$ and $v$ does not have {\rm v}-structure.  
	\end{enumerate} 
	Then, $\mathbf{Z}$ d-separate $\mathbf{X}$ and $\mathbf{Y}$, denoted as $\mathbf{X} \upmodels_\mathcal{G} \mathbf{Y} | \mathbf{Z}$, if $\mathbf{Z}$ blocks any $\mathbf{p}$.  
\end{definition}

Then, two assumptions, causal Markov (or Markov property) and faithfulness, are introduced as follows:
\begin{assumption}[Causal Markov~\cite{BayesianNetwork1985pearl, belief1986pearl, simulation1987pearl, IntelligentSystem1988pearl, IdentifyDuLi1990pearl}]
	Probability distribution $P$ is Markovian to a Bayesian network $\mathcal{G}$ if 
	\begin{equation}
		\mathbf{X} \upmodels_\mathcal{G} \mathbf{Y} | \mathbf{Z} \Rightarrow \mathbf{X} \upmodels \mathbf{Y} | \mathbf{Z},
	\end{equation}
	where $\mathbf{X}$, $\mathbf{Y}$ and $\mathbf{Z}$ are three disjoint subsets.
	\label{assumption: Markov}
\end{assumption}
\begin{assumption}[Faithfulness~\cite{BayesianNetwork1985pearl, belief1986pearl, simulation1987pearl, IntelligentSystem1988pearl, IdentifyDuLi1990pearl}]
	Probability distribution $P$ is faithful to a Bayesian network $\mathcal{G}$ if 
	\begin{equation}
		\mathbf{X} \upmodels \mathbf{Y} | \mathbf{Z} \Rightarrow \mathbf{X} \upmodels_\mathcal{G} \mathbf{Y} | \mathbf{Z}
	\end{equation}
	for all disjoint subsets $\mathbf{X}$, $\mathbf{Y}$ and $\mathbf{Z}$.
	\label{assumption: ff}
\end{assumption}
It is intuitive to understand the two assumptions: the Bayesian network $\mathcal{G}$ corresponds one-to-one to the independence structure of the probability distribution $P$ in the observational data. Thus, if they are satisfied, the joint distribution can be factorized according to the Bayesian network, as follows:
\begin{equation}
	P(X_1, \dots, X_n) = \prod_{i=1}^n P(X_i|\mathbf{Pa}(X_i)), 
\end{equation}
where $\mathbf{Pa}(X_i)$ is the predecessors of $X_i$ in $\mathcal{G}$, that is also called as causal parents.

Moreover, a Bayesian network is also assumed to be causally sufficient, that is
\begin{assumption}[Causal Sufficiency~\cite{BayesianNetwork1985pearl, belief1986pearl, simulation1987pearl, IntelligentSystem1988pearl, IdentifyDuLi1990pearl}]
	Variables $\mathbf{V}$ are said to satisfy causal sufficiency if there is no hidden variable that is a common cause of two or more variables in $\mathbf{V}$.
	\label{assumption: suff}
\end{assumption}
This means that if a Bayesian network (or an SCM) is causally sufficient, there is no hidden path connecting two endogenous variables in $\mathbf{V}$, and all variable information is collected sufficiently to support the network. Actually, the ignorability assumption in POM (see \textbf{Assumption~\ref{assumption: igno}}) also implies the causal sufficiency, and this assumption is weaker than ignorability. But note that, causal sufficiency is hard to satisfy in practice, because many real-world systems (e.g., the economic and climate systems) are complex, thus, we usually cannot collect all the information needed to describe them. Thus, some research works focus on this issue. For example, the Fast Causal Inference algorithm and its variants~\cite{FCI1995, FCI2001, FCI2012, FCI2014, FCI2016, FCI2018} are proposed to discover causality in the presence of hidden variables. And some other related works~\cite{soft2019NIPS, soft2020NIPS, soft2023NIPS} investigate causal discovery with soft intervention, which intervenes in the SCM without altering the network structure. Related works~\cite{OAS2020JMLR, OAS2021NIPS, OAS2021AAAI, OAS2022Biometrika} have also proposed approaches for identifying optimal and efficient adjustment sets, in witch all variables are observable, minimal and valid, and or which removing any variables would invalidate the adjustment.


\section{Differences between POM and SCM}
\label{section: diffs}

In this section, the fundamental differences between POM and SCM are discussed, focusing on three aspects. (i) Counterfactual falsifiability: POM assumes positivity; therefore, counterfactuals in POM are, in principle, supported by observations and can be empirically falsified. In contrast, counterfactuals in SCM may lie outside the support of the observed data and are thus unfalsifiable. (ii) Ignorability and the back-door criterion: POM requires the ignorability assumption to ensure unconfoundedness and the validity of counterfactual inference. In practice, satisfying ignorability typically requires identifying appropriate covariates. This is a complex, human-guided process. By contrast, SCM achieves unconfoundedness through causal sufficiency and back-door adjustment, which are equivalent to the ignorability assumption. SCM facilitates counterfactual inference through graphical language. (iii) Direct causation and indirect causation: POM struggles to process the indirect causation, and it typically relies on the Baron-Kenny model. In contrast, SCM has the advantage of adjusting for the front-door path (indirect causation) by equivalently applying the back-door criterion twice. We further leverage this advantage to design a forward counterfactual inference algorithm that autonomously infers spatio-temporal counterfactuals. The following are the details.

\subsection{Counterfactual Falsifiability}

As shown in Fig.~\ref{figure: ladder}, consider the following query, ``Observed $X_1=0.5, X_2=1, X_3=1.5$, if had done $X_2=2$, what would $X_3 = ?$''. To answer this query, the positivity assumption requires the existence of at least two observational samples that share the same values $X_1=0.5$ but differ in $X_2$. Specially, one with $X_2 = 1$, and another one with $X_2 = 2$ (see \textbf{Assumption~\ref{assumption: posity}}). And then, under the consistency assumption (see \textbf{Assumption~\ref{assumption: consity}}), we can infer counterfactuals by exactly matching the two samples. In contrast to SCM, we can infer the counterfactuals through Pearl's causal ladder, even if we only have a single sample, as shown in Section~\ref{section: pcl}. Thus, how to falsify the counterfactuals? 


This is a challenging problem, as counterfactuals are, by definition, unobserved (see ``?'' in Table~\ref{table: pom}), and we need at least two samples for comparison. In practice, we cannot return to the past to apply a different treatment, nor can we perfectly replicate the same experiment at a different time. Thus, the inferred counterfactuals in SCM cannot be falsified without the assumptions of positivity and consistency, and this view is rejected ambiguously in Pearl's book~\cite{why2018pearl}.

\begin{figure}[htbp]
	\centering
	\begin{minipage}[b]{0.34\textwidth}
		\centering
		\includegraphics[width=\linewidth]{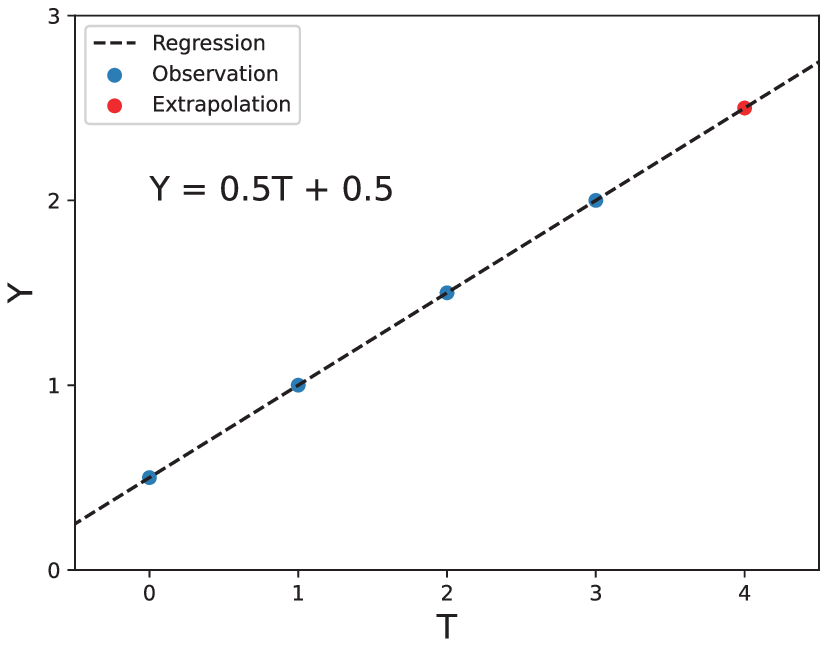}
		\caption{Diagram of data extrapolation.}
		\label{figure: extra}
	\end{minipage}
	\hfill
	\begin{minipage}[b]{0.62\textwidth}
		\centering
		\includegraphics[width=\linewidth]{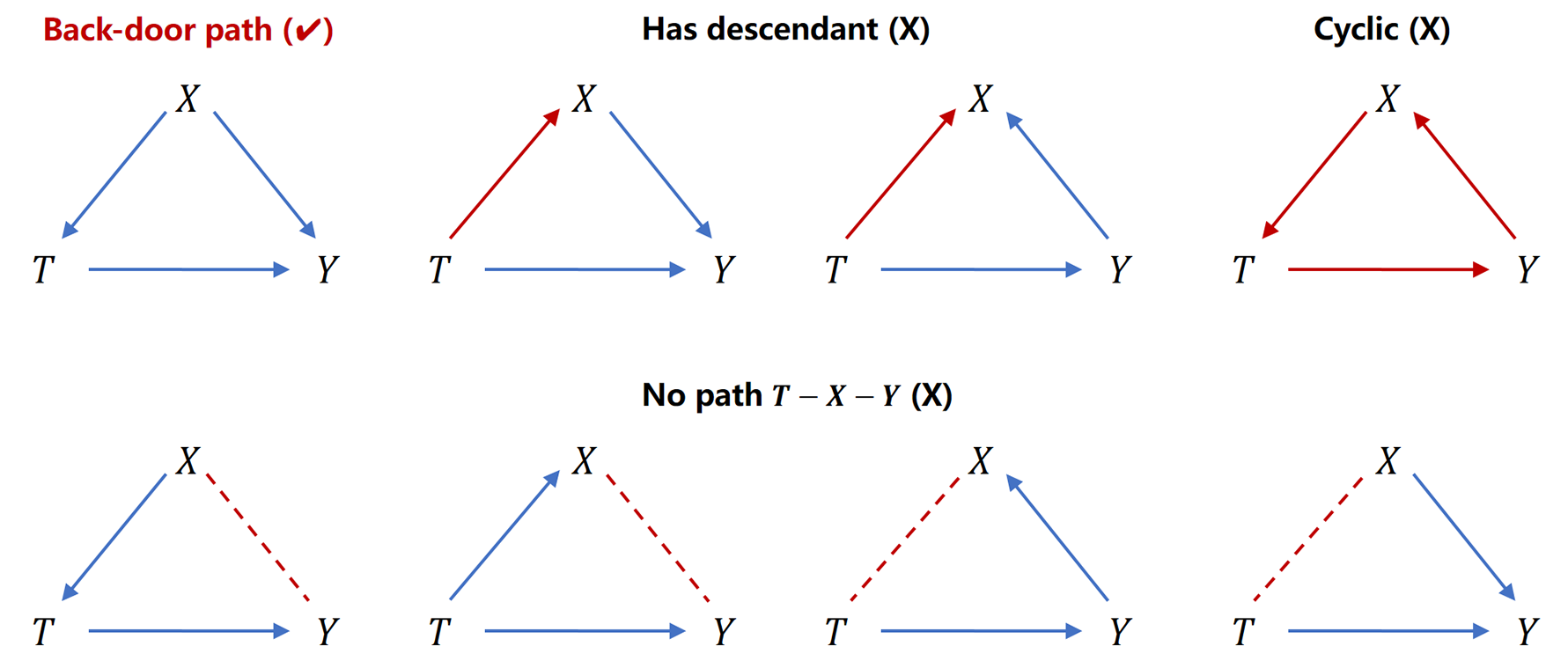}
		\caption{Diagram of a back-door path.}
		\label{figure: backdoor}
	\end{minipage}
\end{figure}

Another explanation is that, without positivity, but with consistency only in POM, counterfactuals can be extrapolated by regressive approaches, and they are falsifiable. For example, as shown in Fig.~\ref{figure: extra}, if there are $T$ and $Y$ only, and the observational data $(T, Y)=\{(0, 0.5), (1, 1), (2, 2.5), (3, 2)\}$ is collected, the linear regressive model can be fitted as $Y=0.5T+0.5$. We then query ``if had done $T=4$, what would $Y=?$''. The counterfactual answer is $Y=2.5$, and the regressive function can falsify the answer because it does not go beyond the observation. Thus, the consistency assumption actually claims that counterfactuals can be falsified by observations. This is also true for SCM, which uses regressive approaches to infer counterfactuals.  


Moreover, there may be ambiguity regarding the preset Bayesian network (see Fig.~\ref{figure: scm}). To build an SCM, a Bayesian network must first be known. However, how to discover a probabilistic network from a single sample? This is usually impossible. A reasonable explanation is that the network structure can be inferred from the system's mechanisms (e.g., physical mechanisms, communication connections) or from graphical knowledge learned from other domains, rather than from probability distribution. However, if it is accepted, the definition of causality could be broader and more ambiguous. This is why probabilistic and causal Bayesian networks are often indistinguishable, as stated in Pearl's book~\cite{why2018pearl}.

Thus, SCM provides a way to think about counterfactuals, emphasizing that the thinking process is counterfactual, not the thinking result, because the result cannot always be falsified without the positivity and consistency assumptions. In contrast, POM emphasizes that the thinking result is counterfactual. This is a fundamental difference in the problem of counterfactual falsifiability.

\subsection{Ignorability and Back-door Criterion}

In the view of statistics, the ignorability assumption (see \textbf{Assumption~\ref{assumption: igno}}) is equivalent in calculation to the causal sufficiency assumption (see \textbf{Assumption~\ref{assumption: suff}}) with a back-door criterion.
The formal definition of the back-door criterion can be found in these papers and books~\cite{diagram1995pearl, overview2009pearl, Causality2009, APrimer2016}. Here we present alternative formulation with respect to Table~\ref{table: pom}, defined as follows:
\begin{definition}[Back-door Criterion~\cite{diagram1995pearl, overview2009pearl, Causality2009, APrimer2016}]
	In Table~\ref{table: pom}, the covariate $X$ is said to satisfy the back-door criterion, if a Bayesian network is built on all the variables relative to $X, Y, T$, and there is an ordered pair $T\rightarrow Y$ that satisfies
	\begin{enumerate}
		\item[1.] no variable in $X$ is a descendant of $T$;
		\item[2.] $X$ blocks every path between $T$ and $Y$ that contains an arrow into $T$.
	\end{enumerate}
	\label{definition: backdoor}
\end{definition}

Thus, for $X, Y, T$, only one structure can satisfy the back-door criterion, as shown in Fig.~\ref{figure: backdoor}. And in the back-door path, it is obtained that
\begin{equation}
	\begin{aligned}
		& P(Y| do(T=1)) \\
		=& \sum_x P(Y| do(T=1), X=x)P(X=x| do(T=1)) \\
		=& \sum_x P(Y| T=1, X=x)P(X=x),
	\end{aligned}
\end{equation} 
which is the same in case of $do(T=0)$~\cite{diagram1995pearl, overview2009pearl, Causality2009, APrimer2016}. Then, if causal sufficiency holds (i.e., no hidden variables confound $X, Y$ and $T$), the expectation of the potential outcomes in Table~\ref{table: pom} can be calculated as 
\begin{equation}
	\begin{aligned}
		&\mathbb{E}[Y| do(T=1)] \\
		=& \sum_y y \times P(Y=y| do(T=1)) \\
		=& \sum_y \sum_x y \times P(Y=y| T=1, X=x)P(X=x) \\
		=& \sum_x P(X=x) \sum_y y \times P(Y=y| T=1, X=x) \\
		=& \mathbb{E}[\mathbb{E}[Y|T=1, X]].
	\end{aligned}
	\label{equation: docal}
\end{equation}
This is the same in POM, as follows:
\begin{equation}
	\begin{aligned}
		& \mathbb{E}[Y(1)] \\
		=& \mathbb{E}[\mathbb{E}[Y(1)| X]]  &/* Law~of~full~expectation*/& \\ 
		=& \mathbb{E}[\mathbb{E}[Y(1)| T=1, X]]  &/* Ignorability*/& \\
		=& \mathbb{E}[\mathbb{E}[Y| T=1, X]],  &/* Consistency*/& \\
	\end{aligned}
	\label{equation: dopom}
\end{equation}
which is the same in case of $do(T=0)$. Thus, with comparing Eq.~(\ref{equation: docal}) and Eq.~(\ref{equation: dopom}), $\mathbb{E}[Y| do(T=1)] = \mathbb{E}[Y(1)]$, and $\mathbb{E}[Y| do(T=0)] = \mathbb{E}[Y(0)]$, in the two frameworks, but they start from different assumptions. Thus, the ignorability assumption can be decomposed as the causal sufficiency assumption and the back-door criterion, and the causal sufficiency is relatively trivial to satisfy. The ignorability is more likely to be a technical assumption that needs many controlled trials to deconfound, while the back-door path is clear to search in a Bayesian network. A potential, more difficult technical issue is accurately discovering the network structure.

\subsection{Direct and Indirect Causation}
\label{section: fci}

\begin{figure}[htbp]
	\centering
	\begin{minipage}[b]{0.57\textwidth}
		\centering
		\includegraphics[width=\linewidth]{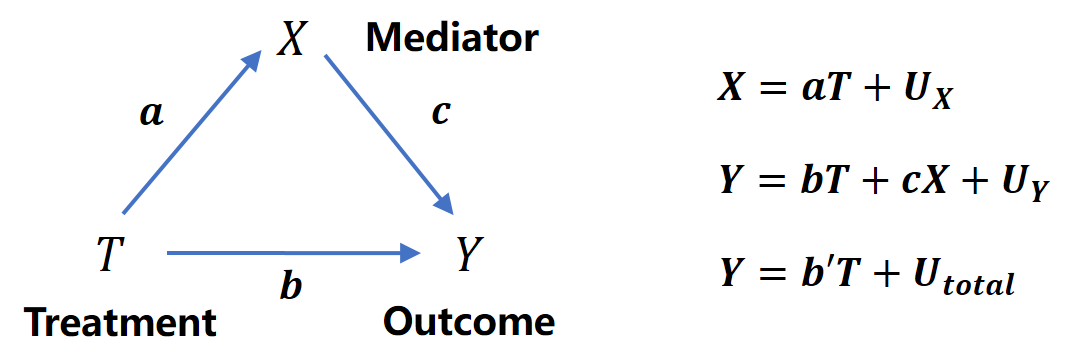}
		\caption{Diagram of mediating analysis~\cite{BK1986}. Here, $a, b, c$ are coefficients and $U_X, U_Y$ are biases. $b'$ and $U_{total}$ are the coefficient and bias, respectively, for the total effect.}
		\label{figure: indirect}
	\end{minipage}
	\hfill
	\begin{minipage}[b]{0.39\textwidth}
		\centering
		\includegraphics[width=0.9\linewidth]{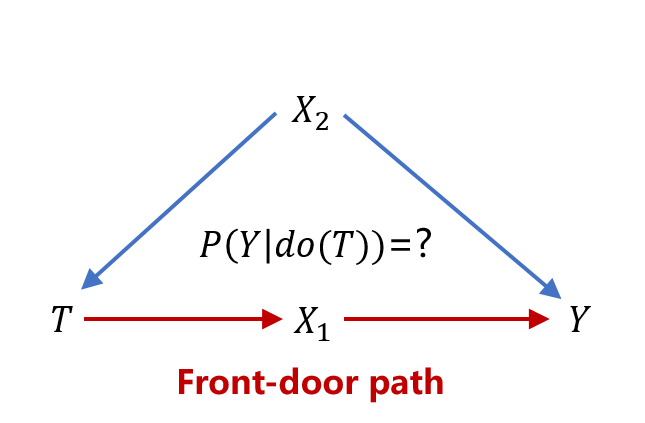}
		\caption{Diagram of front-door adjustment~\cite{APrimer2016}. Here, $T$ is a treatment variable and $Y$ is a target variable. $X_1$ and $X_2$ are both covariates.}
		\label{figure: front_door}
	\end{minipage}
\end{figure}

Suppose that there is a causal ordered pathway from one treatment variable to another outcome variable. Then, if there is at least one variable between the two endpoints, the pathway is called indirect causation, otherwise, it is called directed causation. Indirect causation also called the mediating effect, has been discussed in depth in Rubin's papers~\cite{indirect1997rubin, indirect2003rubin, indirect2004rubin}. Here, the Baron-Kenny model~\cite{BK1986, Mediation2022heyas} is introduced to present a mediating analysis for a potential outcome using a graph. As shown in Fig.~\ref{figure: indirect}, $T\rightarrow Y$ is direct causation, and $T\rightarrow X\rightarrow Y$ is indirect causation.

To analyse indirect causation, the causal-steps approach~\cite{BK1986} is proposed by building three linear regression models, as shown in Fig.~\ref{figure: indirect}. Suppose that the regressive models are fitted sufficiently, the indirect causation would be detected if satisfy
\begin{enumerate}
	\item[1.] coefficients $a, b$ and $c$ are significant;
	\item[2.] $|b| < |b'|$.
\end{enumerate}
Another approaches are to test the significance of $H_0: b' - b = 0$~\cite{coefficient1993mackinnon, coefficient2004mackinnon, coefficient2007mackinnon}. Or, to test the significance of $H_0: ab=0$~\cite{sobel1982, sobel1986}. If the correct regressive models are built, counterfactuals can be calculated, just as in SCM in Fig.~\ref{figure: scm}. Moreover, they are all regression-based approaches that are limited for use in large multivariate datasets, because one needs to test whether a covariate is a confounder or a mediator, and this process would be seriously time-consuming.

However, the test process can be faster with a Bayesian network in an SCM. Here, we propose an algorithm for forward counterfactual inference, as shown in \textbf{Algorithm~\ref{algorithm: FCI}}. To clearly demonstrate the principle of \textbf{Algorithm~\ref{algorithm: FCI}}, we first introduce the front-door criterion as follows:
\begin{definition}[Front-door Criterion~\cite{diagram1995pearl, overview2009pearl, Causality2009, APrimer2016}]
	The covariate $X$ is said to satisfy the front-door criterion, if a Bayesian network is built on all the variables relative to $X, Y, T$, and there is an ordered pair $T\rightarrow Y$ that satisfies 
	\begin{enumerate}
		\item[1.] $X$ intercepts all directed paths from $T$ to $Y$;
		\item[2.] there is no back-door path from $T$ to $X$;
		\item[3.] all back-door paths from $X$ to $Y$ are blocked by $T$.
	\end{enumerate}
\end{definition}

\begin{figure*}[htbp]
	\centering
	\includegraphics[width=\linewidth]{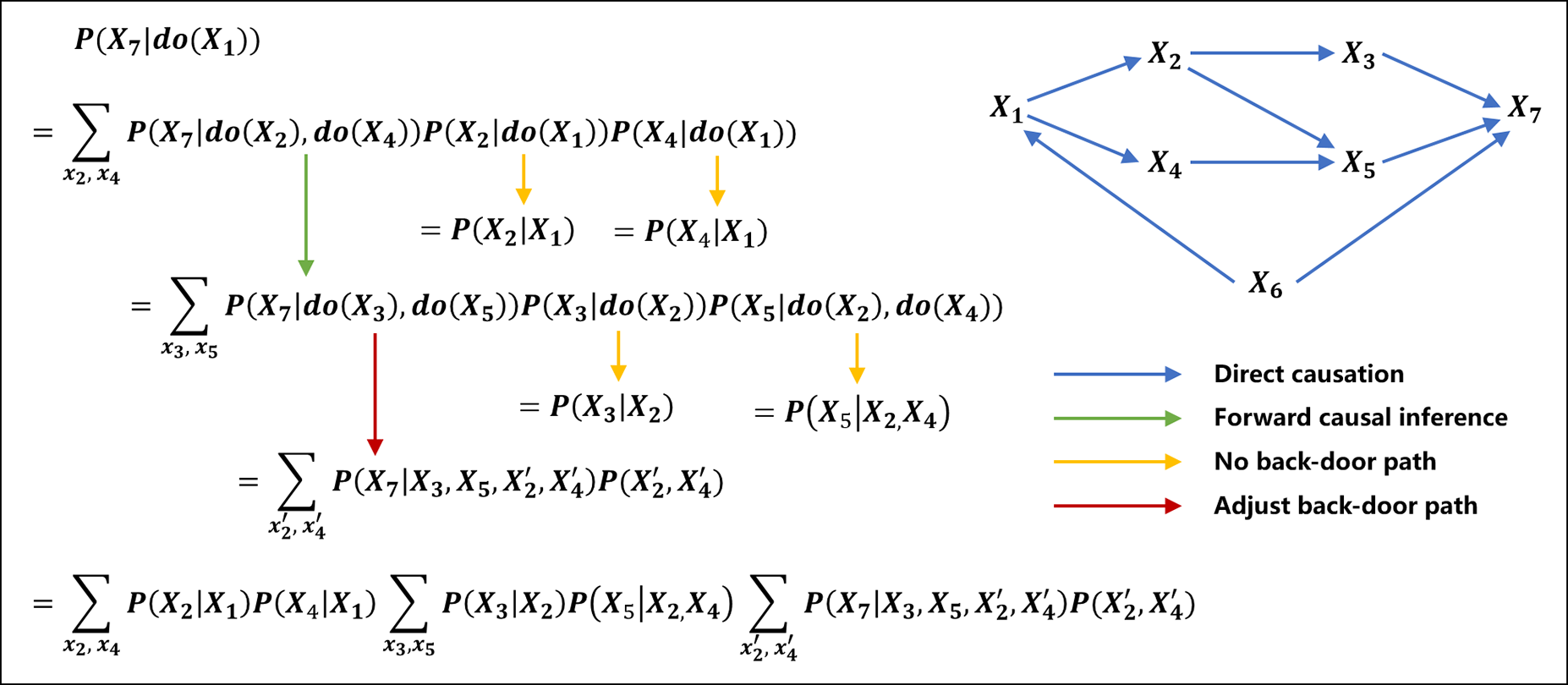}
	\caption{Example of the Forward Counterfactual Inference algorithm. Here, $X_i$ is the same as $X_i'$ to distinguish between the two recursions.}
	\label{figure: fci}
\end{figure*}

Then, a classic example is provided to show the front-door adjustment using Theorem~3.4.1 in book~\cite{APrimer2016}, as shown in Fig.~\ref{figure: front_door} here. If given a treatment variable $T$ and a target variable $Y$, with two covariates $X_1$ and $X_2$. Then one can adjust the distribution $P(Y| do(T))$ with the front-door criterion as follows:
\begin{equation}
	\begin{aligned}
		P(Y| do(T)) &= \sum_{x_1} P(X_1 | do(T)) P(Y | do(X_1)) \\ 
		&= \sum_{x_1} P(X_1 | T) \sum_{t'} P(Y | X_1, T') P(T') 
	\end{aligned}
	\label{equation: front_door}
\end{equation}
Thus, we can conclude from this equation:
\begin{enumerate}
	\item[1.] If there is one or multiple ordered causal pathways from $T$ to $Y$ and there is at least one mediator on the way, then the causal pathway can be decomposed into two causal pathways through that mediator, e.g., $T\rightarrow X_1$ and $X_1 \rightarrow Y$ in Eq.~(\ref{equation: front_door}). 
	
	\item[2.] If there is no mediator on all causal pathways, then adjustment using back-door criterion. Of course, if there is no back-door path, no need to adjust. 
\end{enumerate}
Thus, the front-door principle is non-atomic and decomposable. With this more foundational principle in mind, we propose the forward counterfactual algorithm to autonomously apply it, as presented in \textbf{Algorithm~\ref{algorithm: FCI}}. The algorithm can be applied to an arbitrary set of variables $\{X_1, \dots, X_n\}$ that satisfy the causal sufficiency condition. We also provide an example to illustrate the inference process, as shown in Fig.~\ref{figure: fci}.

\begin{algorithm}[htbp]
	\caption{Forward Counterfactual Inference}
	\label{algorithm: FCI}
	\LinesNumbered
	\KwIn{A Bayesian network $\mathcal{G}$ built on $\mathbf{V}=\{X_1, \dots, X_n\}$ with one target $X_i$ and multiple sources $X_{j_1}, \dots, X_{j_p}$.}
	\KwOut{Distribution $P(X_i|do(X_{j_1}), \dots, do(X_{j_p}))$.}
	
	Search one or multiple ordered causal pathways in $\mathcal{G}$ from the sources $X_{j_1}, \dots, X_{j_p}$ to the target $X_i$; 
	
	\If{$X_{j_1}, \dots, X_{j_p}$ are all the direct causation for $X_i$}{
		Search back-door paths (common causes) starting from $X_{j_1}, \dots, X_{j_p}$ and $X_i$ reversely in $\mathcal{G}$. \\
		\If{no back-door path within them}{
			\Return $P(X_i|X_{j_1}, \dots, X_{j_p})$;
		}
		\Else{
			For every back-door path, select a group of direct predecessors, $\mathbf{W}\subset \mathbf{V}$, and then, according to the back-door criterion (see \textbf{Definition~\ref{definition: backdoor}}), \\
			\Return $\sum_w P(X_i|X_{j_1}, \dots, X_{j_p}, W)P(W)$;
		}  
	}
	
	Otherwise, for $q$ pathways starting from $X_{j_p}$, select $q$ first mediators $X_{j_{p_1}}, \dots, X_{j_{p_q}}$. Then, a group of source-mediator pairs, e.g., $(X_{j_p}, X_{j_{p_q}})$, can be obtained;
	
	Factorize $P(X_i|do(X_{j_1}), \dots, do(X_{j_p}))$ as 
	\begin{equation}
		\begin{aligned}
			&P(X_i|do(X_{j_1}), \dots, do(X_{j_p})) \\
			=& \sum_{x_{j_{1_1}: j_{p_q}}} P(X_i|do(X_{j_{1_1}}), \dots, do(X_{j_{2_1}}), \dots, do(X_{j_{p_q}})) \\
			&\times \prod_p \prod_q P(X_{j_{p_q}}|do(X), X\in \mathbf{Pa}(X_{j_{p_q}})\subseteq \{X_{j_1: j_p}\}),
		\end{aligned}
		\label{equation: fact}
	\end{equation}
	where $\mathbf{Pa}(X_{j_{p_q}})$ is the causal parents of $X_{j_{p_q}}$;
	
	Solve each distribution $P(\cdot| do(\cdot), \dots, do(\cdot))$ in Eq.~(\ref{equation: fact}) by \textbf{Algorithm~\ref{algorithm: FCI}} recursively.
	
	\Return $P(X_i|do(X_{j_1}), \dots, do(X_{j_p}))$.
\end{algorithm}

In \textbf{Algorithm~\ref{algorithm: FCI}}, the approaches to realize the codes in lines 1 and 3 can be various, e.g., some variants of the depth-first-search algorithm and breadth-first-search algorithm~\cite{algorithms2022mit}. Thus, their time complexities are both $\mathcal{O}(|\mathbf{V}| + |\mathbf{E}|)$, if a Bayesian network $\mathcal{G}=(\mathbf{V}, \mathbf{E})$ is given, where $\mathbf{V}$ is the node set and $\mathbf{E}$ is the edge set. Meanwhile, if the computational cost of factorization and adjustment are overlooked, the time complexity of the code in line 9 in \textbf{Algorithm~\ref{algorithm: FCI}} is $\mathcal{O}(|\mathbf{V}|)$, because we only need to traverse all mediators in each pathway forward, and this process can be performed in parallel. Thus, the total time complexity of \textbf{Algorithm~\ref{algorithm: FCI}} is $\mathcal{O}(|\mathbf{V}|^2 + |\mathbf{V}||\mathbf{E}|)$. Thus, if the network structure of the Bayesian network is large and can be discovered accurately, the efficiency of SCM to infer counterfactuals would be better with polynomial time complexity.

\section{Spatio-Temporal Graphical Counterfactuals}
\label{section: stgc}

Then, for the examples of counterfactual questions at the beginning of this article, that is, if a different investment strategy had been implemented, would we obtain higher returns? And, in computer networks, what changes would occur in network load if a node's configuration had never been changed? Would an individual still purchase the product had they not been exposed to the advertisement? In these real-world scenarios, interactive behaviors with lagged causal effects exist widely. POM cannot answer these questions, because it does not allow the interactive behaviors between experimental units (see SUTVA in \textbf{Assumption~\ref{assumption: sutva}}), even with some temporal quasi-experimental approaches (e.g., Regression Discontinuity Design~\cite{RDD2001, RDD2008} and Differences in Differences~\cite{DiD2004, DiD2011}). This applies to SCM as well, because the foundational Bayesian network is a DAG that also does not allow mutual connections between network nodes. Thus, to answer these counterfactual queries for interactive units and lagged causal effects, we propose a concept of spatio-temporal graphical counterfactuals.

\subsection{Spatio-Temporal Bayesian Networks}

\begin{figure*}[htbp]
	\centering
	\includegraphics[width=\linewidth]{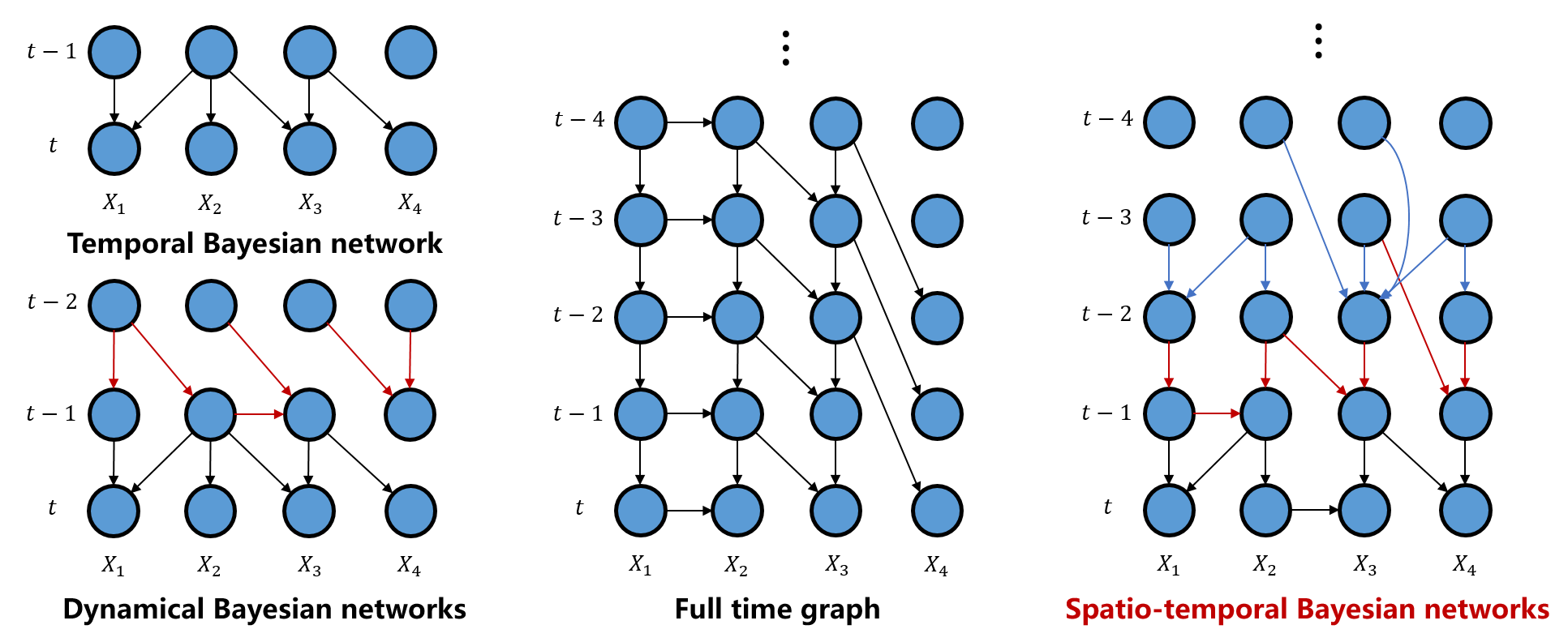}
	\caption{Diagrams of various causal graphs.}
	\label{figure: graphs}
\end{figure*}

Before the SCM, various causal graphical models have been proposed to model the spatio-temporal causality. As shown in Fig.~\ref{figure: graphs}, a temporal Bayesian network~\cite{PGM2009, OCE2015SIAM} is proposed to model the momentary causal dependency in a first-order Markov process. Further, dynamical Bayesian networks~\cite{PGM2009} are proposed to represent a combination of multiple temporal Bayesian networks. Full time graph~\cite{Elements2017} is proposed to summarize temporal Bayesian dependencies with complete timestamps. Full time graph allows the instantaneous causal effects between two units (or variables) in a high-order Markov process, but it only supports strictly stationary causality, that is, causal dependencies that do not change over time. Moreover, the full time graph can be uniquely decomposed into multiple high-order temporal Bayesian networks, named UCN, if the instantaneous causal effects are not allowed~\cite{UCN2022}. The comparisons of these causal graphs are shown in Table~\ref{table: graphs}.

\begin{table*}[htbp]
	\centering
	\caption{Comparisons of various causal graphs.}
	\label{table: graphs}
	\renewcommand\arraystretch{1.3}
	\setlength{\tabcolsep}{1.7mm}
	\small
	\begin{tabular}{ccccc} 
		\hline 
		& \makecell{Temporal Bayesian \\ network} & \makecell{Dynamical Bayesian \\ networks} & Full time graph & \makecell{Spatio-temporal \\ Bayesian networks} \\
		\hline  
		High-order causation & \textbf{$\times$} & \textbf{$\times$} & $\checkmark$ & $\checkmark$\\
		Instantaneous causation & \textbf{$\times$} & $\checkmark$ & $\checkmark$ & $\checkmark$ \\
		Nonstationary causation & \textbf{$\times$} &  $\checkmark$ & \textbf{$\times$} & $\checkmark$ \\
		\hline 
	\end{tabular}
\end{table*}

Thus, based on the characteristics of these causal graphs, STBNs are defined as a group of DAGs indexed by different timestamps, as shown in Fig.~\ref{figure: graphs}. In STBNs, nodes are not equivalent to variables. They represent the state of the variables at some time steps. Under the assumptions of causal Markov (see \textbf{Assumption~\ref{assumption: Markov}}) and faithfulness (see \textbf{Assumption~\ref{assumption: ff}}), the joint distribution of all nodes can be factorized as 
\begin{equation}
	P(X_{1, t}, \dots, X_{n, t}, X_{1, t-1}, \dots, X_{n, t-T+1}) = \prod_{\tau=0}^{T-1} \prod_{i=1}^{n} P(X_{i, t-\tau}| \mathbf{Pa}(X_{i, t-\tau})),
\end{equation}
where $X_1, \dots, X_n$ are the variables, and the timestamps are in the range of $0\sim T-1$. Theoretically, STBNs allow an infinite number of time steps, but in practice, they are typically finite.

To guarantee the global and unique DAG in STBNs, a temporal assumption is introduced as
\begin{assumption}[Temporal Assumption]
	In STBNs, the cause node precedes or occurs concurrently with the effect node.
\end{assumption}
This assumption does not allow the directed cause-effect pairs from now to the past, e.g., $X_{i, t-\tau_1}\rightarrow X_{j, t-\tau_2}, \tau_1 < \tau_2$. And this guarantees a functionally identifiable network structure without Markov equivalence classes~\cite{Elements2017, UCN2022, identifiability2008nips}. Moreover, the causal sufficiency assumption (see \textbf{Assumption~\ref{assumption: suff}}) is also needed to guarantee the unconfoundedness. 

Then, if STBNs are given, the spatial-temporal counterfactuals can be inferred through \textbf{Algorithm~\ref{algorithm: FCI}}, because the global and local STBNs are both DAGs. Compared with naive Bayesian networks, STBNs incorporate spatio-temporal concepts, and the inference process naturally proceeds forward in time. This enables momentary interventions at every time step, and all interventions work sequentially rather than simultaneously, as in \textbf{Algorithm~\ref{algorithm: FCI}}. If this property is considered thoroughly, the inference algorithm can be faster.


\subsection{Nonstationarity of STBNs}

\begin{figure}[htbp]
	\centering
	\includegraphics[width=0.76\linewidth]{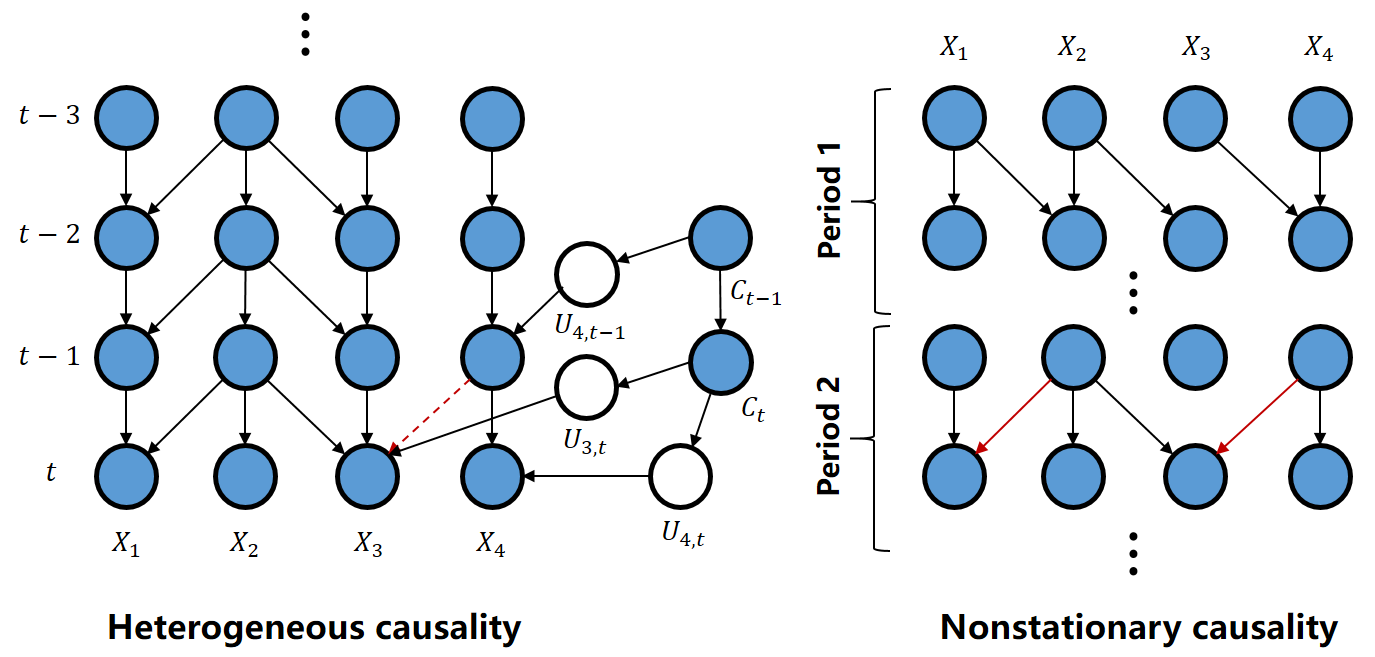}
	\caption{Diagrams of nonstationary causality and heterogeneous causality. Here, $C_{t-\tau} = t-\tau, \tau=0, \dots, T-1$ are usually assumed; Thus, they are observable, instead of a hidden variable.}
	\label{figure: nonstt}
\end{figure}

Nonstationarity is a significant topic in researches on time series. However, the reasons for it are likely to be diff from the view of causality. Three of the most popular reasons are:
\begin{enumerate}
	\item[1.] \textbf{Trending:} The time series has ascending or descending trending for long terms.  
	\item[2.] \textbf{Time-varying Variance:} The variances of the additive noise are time-varying, a phenomenon often referred to as heterogeneous causality in the literature~\cite{nonstationary2017kun, nonstationary2020huang}.
	\item[3.] \textbf{Nonstationary Causality:} The structures of STBNs change over time, and at least once, as shown in Fig.~\ref{figure: nonstt}. 
\end{enumerate}

For trending time series, an effective approach is to calculate the differences of the time series during preprocessing. This approach obtains the differential time series, that is, the observations of the system dynamics, as in Eq.~(\ref{equation: netdy}). This implies that the dynamics is stable over the period, and that the structures of STBNs do not change, that is, the differential time series are stationary. Thus, if the dynamics equations, e.g., Eq.~(\ref{equation: netdy}), are not equal to zero, the causal links in STBNs are identifiable through the Jacobian matrix of dynamics~\cite{universality2013barzel, differential2020NIPS, differential2021pengcui, differential2023anzhang}.

In the case of time-varying variance, causal identifiability relies on the mutual independence of exogenous additive noises~\cite{identifiability2008nips}. Specially, if no causal connection exists between two additive noise nodes, the causation is identifiable. Thus, an effective approach is to relate the time-varying noises with a common timestamp, treating the timestamps at each time step as observable confounders $C_{t-\tau}, \tau=0, \dots, T-1$~\cite{nonstationary2015huang, nonstationary2017kun, nonstationary2019huang, nonstationary2020huang}. As shown in Fig.~\ref{figure: nonstt}, there are three causal pathways, that is, $C_{t-1}\rightarrow U_{4, t-1}\rightarrow X_{4, t-1}$, $C_{t-1}\rightarrow C_{t}\rightarrow U_{3, t}\rightarrow X_{3,t}$, and $C_{t-1}\rightarrow C_{t}\rightarrow U_{4, t}\rightarrow X_{4,t}$. Thus, if $C_{t}$ and $C_{t-1}$ are hidden, the spurious connection between $X_{4, t-1}$ and $X_{3,t}$ would be detected, like the dotted line in Fig.~\ref{figure: nonstt}. However, $C_{t}$ and $C_{t-1}$ are usually known as the timestamp $t$ and $t-1$, thus, $C_{t}$ and $C_{t-1}$ block the fork paths between $X_{4, t-1}, X_{3,t}$ and $X_{4,t}$. Then, the noise variables $U_{4, t-1}, U_{3, t}$ and $U_{4, t}$ are mutually independent. Thus, causal identifiability is preserved. Note that, however, the premise to do so is that any hidden confounders associated with time-varying variance can be expressed as smooth or non-smooth functions of timestamps~\cite{nonstationary2020huang}.

Nonstationary causality presents a significant challenge, particularly in scenarios involving frequently changing structures. To identify the nonstationary causality, one approach is to use sliding windows to identify causation within different time periods, as shown in Fig.~\ref{figure: nonstt}. For each sliding window, it is assumed that the causal structure is invariant, and then, a group of samples would be input to identification algorithms, e.g.,~\cite{nonstationary2016arxiv, Invariant2016, Invariant2018, dynotears2020, UCN2022, NHCE2024}, to identify an invariant structure during this period. However, if network changes are frequent, the samples used for identification over a short period would be few. In contrast, we usually need sufficient samples to assess the significance of causal connections. Thus, another reasonable approach is to first identify the change points, and then, determine the causality for the corresponding periods~\cite{changepoint2007, changepointi2017}. However, this still does not work in the extreme case of frequently changing, because the samples between two change points may be few. Moreover, there are other approaches that view nonstationary causality as a probabilistic normalized flow, and then use variational Bayesian inference to estimate it globally~\cite{normalization2015ICML, normalization2021JMLR, normalization2023NIPS}.

\subsection{STBNs and Complex networks}

\begin{figure}[htbp]
	\centering
	\includegraphics[width=0.7\linewidth]{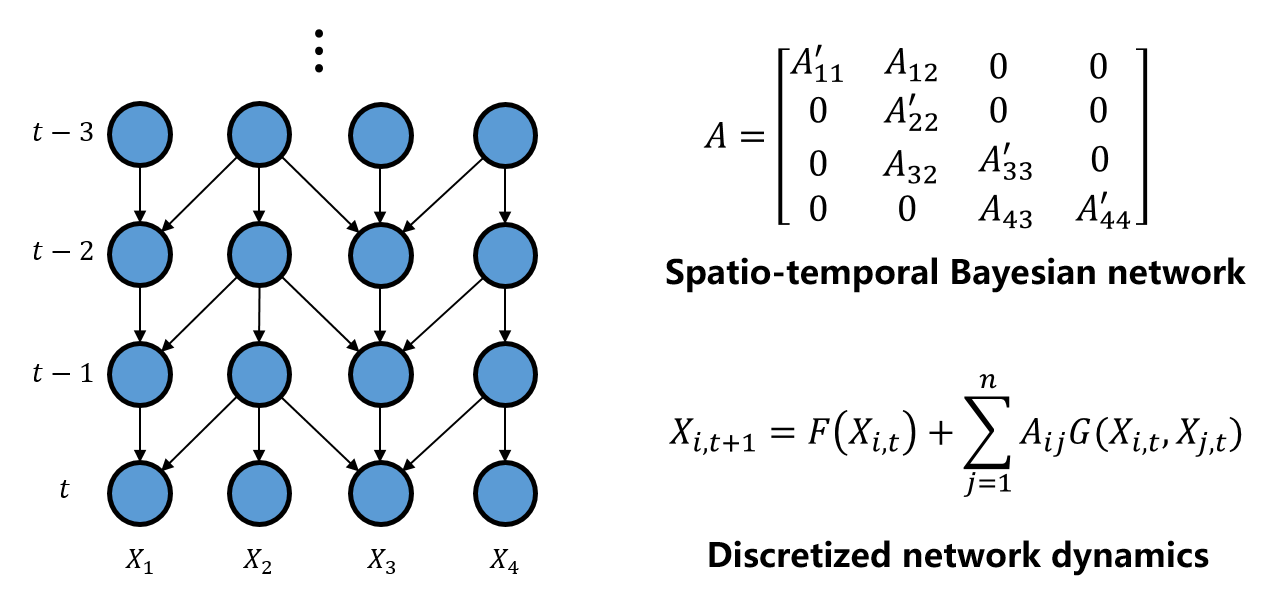}
	\caption{Diagram of discretized network dynamics. Here, $A$ is the adjacency matrix of the first-order STBNs, and $A_{ii}', i=1,\dots, n$ represents the joined causal effect of $F$ and $G$.}
	\label{figure: netdy}
\end{figure}

Here, we suppose STBNs model has been built, and it is associated with an interactive time-series process of multivariate (or multiple units). Then, we introduce a complex network, another network model, to examine the causal interactions in a complex system. The network dynamics for variables $X_i, \dots, X_n$ can be defined as
\begin{equation}
	\frac{dX_{i, t}}{dt} = F(X_{i, t}) + \sum_{j=1}^n A_{ij} G(X_{i, t}, X_{j, t}),
	\label{equation: netdy}
\end{equation} 
where $F(\cdot)$ is the self-governed function of $X_i$, and $G(\cdot, \cdot)$ is the interactive function for node pairs $X_i$ and $X_j$. Moreover, $A_{ij} \geq 0$ is the connection between $X_i$ and $X_j$. Thus, if Eq.~(\ref{equation: netdy}) is discretized, we can find it the same as first-order STBNs on stationary time series, as shown in Fig.~\ref{figure: netdy}.

First-order STBNs exhibit advantageous properties when the underlying network dynamics process certain structural features. For example, determining how to affect a node in STBNs by ``intervening'' another node, the practical inference steps may not be long, due to the small-world property~\cite{smallworld1998, smallworld1999newman} or the scale-free property~\cite{scalefree1999} of complex networks. Moreover, counterfactual outcomes would not change because network systems have tolerance from external ``interventions''~\cite{tolerence2000, robustness2011}. We would also know the synchronization~\cite{synchinization2002, synchinization2008, synchinization2016, synchinization2022}, controllability~\cite{controllability2002guanrongchen, controllability2005jinhulv, controllability2009wenwuyu, controllability2012wenwuyu, controllability2013wenwuyu, controllability2013wenwuyu2, controllability2011, temporal2017amingli}, resilience~\cite{resilience2016jianxigao, resilience2022jianxigao} of STBNs. We would also be able to sparsely identify the network structure even if the network size is large~\cite{identification2022tpi, identification2022tsp}, because the independence is equivalent to predictability in STBNs~\cite{UCN2022}. Also, the network size of STBNs can be reduced, because many complex networks can be simplified and still provide an insightful description of the causality of interest~\cite{simplicity2024jianxigao, simplicity2024lowrank}. This reduction supports the design of a counterfactual inference algorithm on smaller networks, thereby improving inference efficiency.

\section{Conclusion}

This work focuses on the spatio-temporal graphical counterfactuals and presents an overview of their theoretical foundations and application approaches. To discuss the theoretical basis, a survey is conducted, and the definition of counterfactuals is based on the concept of potential outcomes in the POM framework. We then, introduce the SCM framework, which infers counterfactuals via graphical languages equivalent to those in POM. Further, to infer counterfactuals on intelligent machine autonomously, a Forward Counterfactual Inference algorithm is designed in this work, and it can recursively solve the counterfactual probability distribution, $P(X_i|do(X_{j_1}), \dots, do(X_{j_p}))$, with polynomial time complexity if multi-nodes interventions are conducted on $X_{j_1}, \dots, X_{j_p}$. With this algorithm, spatio-temporal graphical counterfactuals can be inferred, given two elements: the network structure of STBNs and the algorithm identifying it. This work discusses structure identification algorithms under various nonstationary conditions and examines the feasibility of improving algorithm efficiency from the perspective of complex networks.

\Acknowledgements{This work was supported in part by the National Key R\&D Program of China (Grant No. 2022ZD0120004), the Youth Scientist Project of the Ministry of Science and Technology of China (Grant No. 2025YFF0524100), the National Natural Science Foundation of China (Grant Nos. 62233004, 62273090, 62073076, T2541017, 52525204, 52572354), the Zhishan Youth Scholar Program of the Southeast University, the Jiangsu Provincial Scientific Research Center of Applied Mathematics (Grant No. BK20233002), the basic research program of Jiangsu  (Grant No. BK20253020), the Open Research Project of the State Key Laboratory of Industrial Control Technology, China (Grant No. ICT2025B54), the US National Science Foundation under (Grant No. 2047488) and the Rensselaer-IBM AI Research Collaboration.}

\end{document}